\documentclass[journal,onecolumn]{IEEEtran}
\usepackage{cite}
\usepackage{amsmath}
\usepackage{graphicx}
\usepackage{algorithm}
\usepackage{algpseudocode}
\usepackage{multirow}
\usepackage{subcaption}
\usepackage[table,xcdraw]{xcolor}
\usepackage{authblk}

\algblock{Input}{EndInput}
\algnotext{EndInput}
\algblock{Output}{EndOutput}
\algnotext{EndOutput}

\newcommand*{\affaddr}[1]{#1} 
\newcommand*{\affmark}[1][*]{\textsuperscript{#1}}
\newcommand*{\email}[1]{\textit{#1}}

\begin{document}

\title{Fake Reviews Detection through Analysis of Linguistic Features}


\author{%
Faranak Abri\affmark[1], Luis Felipe Gutiérrez\affmark[1], Akbar Siami Namin\affmark[1], Keith S. Jones\affmark[2], and David R. W. Sears\affmark[3]\\
\affaddr{\affmark[1]Department of Computer Science,}
\affaddr{\affmark[2]Department of Psychological Sciences,}
\affaddr{\affmark[3]Performing Arts Research Lab}\\
\affaddr{\affmark[1,2,3]Texas Tech University}\\
\email{\{faranak.abri, luis.gutierrez-Espinoza, akbar.namin, keith.s.jones, david.sears\}@ttu.edu}\\
}

\maketitle

\begin{abstract}
    Online reviews play an integral part for success or failure of businesses. Prior to purchasing services or goods, customers first review the online comments submitted by previous customers. However, it is possible to superficially boost or hinder some businesses through posting counterfeit and fake reviews. This paper explores a natural language processing approach to identify fake reviews. We present a detailed analysis of linguistic features for distinguishing fake and trustworthy online reviews. We study 15 linguistic features and measure their significance and importance towards the classification schemes employed in this study. Our results indicate that fake reviews tend to include more redundant terms and pauses, and generally contain longer sentences. The application of several machine learning classification algorithms revealed that we were able to discriminate fake from real reviews with high accuracy using these linguistic features\footnote{This paper is the pre-print of a paper to appear in the proceedings of the IEEE International Conference on Machine Learning Applications (ICMLA'20) entitled:``{\it Linguistic Features for Detecting Fake Reviews.''}}. 
\end{abstract}

\begin{IEEEkeywords}
    fake review, deception detection, machine learning, linguistic features
\end{IEEEkeywords}

\section{Introduction}
\label{sec:intro}

Detecting lies and deception has been a long-standing research problem. With its root in courtrooms, where it is necessary for judges to differentiate between lies and truths, the lie detection problem has been studied extensively by interdisciplinary researchers from academia and practitioners from law enforcement and police departments. For instance, the SCAN (Scientific Content Analysis) technique, which was developed by Sapir~\cite{Sapir}, a police lieutenant who served as a polygraph examiner. The technique uses the words criminal suspects use to reason whether what they said is accurate. The effectiveness of SCAN has been discussed in several experimental studies~\cite{Smith2001}. Similarly, there have been several studies on developing technologies for the purpose of automatically detecting lies~\cite{Zhou2003, Zhou2004}. These data-driven techniques often utilize personal traits and features to decide whether a person is deceptive. 

Automated deception and lie detection techniques are usually grouped into four approaches: those that analyze 1) transcribed speech features \cite{Zhou2003, Zhou2004}, 2) audio and speech features \cite{Eurospeech2005}, 3) video and image features \cite{ICML2015, AAAI2018}, and 4) hybrid methodologies \cite{Interspeech2017}. This paper focuses on the analysis of transcribed textual features in the context of differentiating fake from real online reviews. 

When transcribed speech is analyzed, the focus is usually on linguistic features, including~\cite{Zhou2003, Zhou2004}:

\begin{itemize}
    \item {\it Quantity} including average number of words, verbs, modifiers, noun phrases, and sentences in the speech.
    \item {\it Complexity} including average number of clauses, sentence/word length, noun phrase length, and pausality.
    \item {\it Non-Immediacy} including passive voice, modal verb, objectification(i.e., replicable), uncertainty, generalizing terms, self and group reference, etc.
    \item {\it Expressiveness/Emotiveness} (i.e., $\frac{\#Adjectives+\#Adverbs}{\#Nouns+\#Verbs}$).
    \item {\it Diversity} including lexical diversity (i.e., \% unique words), content word diversity
(i.e., \% unique content words), redundancy (i.e., \% function words).
    \item {\it Informality} including typo ratio: $\frac{\#Typo}{\#words}$
    \item {\it Specificity} including spatio-temporal information, perceptual information, positive and
negative affect.
\end{itemize}

While the effectiveness of these linguistic cues has been studied in transcribed speech data, their performance in the context of textual data to identify fake and counterfeit reviews has not been explored yet. Unlike transcribed speech data, online reviewers have plenty of time to think about their deceitful or trustworthy thoughts and thus devise more convincing textual reviews or fake statements and reviews. This freedom in choosing the correct wordings causes daunting challenges when distinguishing lies and truths through linguistic features in online settings. 

Furthermore, the significance level of these linguistic features towards the accuracy of the underlying classification problem and technique is unknown. It is desirable to reduce the number of linguistic features utilized in the classification problem mainly to 1) simplify the classification model, 2) reduce the amount of noise and uncertainty introduced by less important features, and 3) ignore correlated features and their possible interactions in classification modeling. 

Aligned with the aforementioned challenges and problems, this paper implements the above linguistic features~\cite{Zhou2003, Zhou2004} in order to investigate the effectiveness of these features in enabling detection of fake online reviews. More specifically, we apply feature selection and reduction techniques, such as Recursive Feature Elimination (RFE) \cite{guyon2002gene}, and random forest to identify the key linguistic features and their significance along with their ability to distinguish between fake and real reviews. This paper makes the following key contributions:
\begin{itemize}
    \item It introduces a new dataset, called the Restaurant Data Set, to support this line of research.
    \item It investigates the effectiveness of linguistic cues in the context of fake reviews detection.
    \item It measures the significance and interactions between these linguistic cues and identifies less important features. 
\end{itemize}

The rest of this paper is organized as follows: Section \ref{sec:relatedwork} reviews the related work. In Section \ref{sec:background}, we provide the technical background of the classifiers studied in this work. Section \ref{sec:setup} presents the experimental setup. The results of the experiments are reported in Section \ref{sec:results}. Section \ref{sec:discussion} compares our results with the existing work. Section \ref{sec:conclusion} concludes the paper and highlights future research directions.


\section{Related work}
\label{sec:relatedwork}


{\it Features:} Hussain et al.~\cite{Hussain2019} in their survey state that the most applied features for spam review detection are linguistic features. 
They also list several other features, including content of the review, meta-data of the review, information about a product and spammer's behavioral information. Crawford et al.~\cite{Crawford2015} divided the feature types into 1) ``review-centric'', which uses the information in the content of a single review, and 2) ``reviewer-centric'',  which uses review metadata related to the reviewers, such as the information about a specific reviewer or the common information found in all reviews from the same person. 

Generally, linguistic approaches for feature extraction consist of several steps: i) text preprocessing, such as removing extra words or punctuation, Part of Speech (POS) tagging,  and stemming (i.e., converting different forms of a word into a single format), ii) tokenization, which is also called creating uni-gram or bi-gram or more generally n-gram, iii) transformation,  that is, the process of forming a sparse matrix, which shows the frequency of repetition of tokens and can be done by ``simple count'' or ``term frequency and inverse document frequency'' (TF-IDF), and iv) feature selection, which is done to reduce the number of features in order to eliminate insignificant features and improve detection performance~\cite{Hussain2019}. 

{\it Detection:} Hussain et al.~\cite{Hussain2019} identify two main approaches for fake review detection: 1) machine learning, including supervised and unsupervised clustering techniques, and 2) lexicon-based methods, including dictionary-based and corpus-based methods. In addition to these two major approaches, Crawford et al.~\cite{Crawford2015} discussed semi-supervised techniques for machine learning approaches. 

Jindal and Liu~\cite{jindal2007} collected their dataset by crawling reviews from Amazon's Website. The dataset consists of four types of products: books, music, DVDs, and manufactured products. By using three rules, they considered these reviews as spam: 1) duplicate reviews (i.e., reviews with similarity more than 90\%), 2) reviews complementing only about the brand, and 3) reviews irrelevant to the product. They extracted 36 features including 21 review-centric features, such as number of feedbacks and length of the review body, 11 reviewer-centric features, such as average rating by the reviewer, and 4 product-centric features, such as price of the product. They performed logistic regression, Support Vector Machine (SVM), decision tree, and naive Bayesian classification for detecting spam reviews 
and achieved their best result by logistic regression using all the features, which achieved 78\% Area Under the ROC Curve (AUC).

Ott et al.~\cite{2011Ott} created their dataset, which included 400 real reviews and 400 deceptive reviews about hotels. The authors employed anonymous online workers (i.e., Turkers) to create the deceptive reviews, and extracted reviews from legitimate reviewers on TripAdvisor to serve as real reviews. They evaluated their automated algorithms and results against human judgment regarding whether reviews were fake or real. It was shown that human judgment was around random guess. They trained their classifiers using three groups of features: POS, LIWC (Linguistic Inquiry and Word Count)~\cite{LIWC2001}, uni-gram, bi-gram, tri-gram, and also the combination of these features. They performed classifications using linear SVM and Naïve Bayes models. Their best result was achieved using LIWC accompanied by the bi-gram features and SVM model, which achieved 89.8\% accuracy.

Shojaee et al.~\cite{2013Shojaee} examined writing-style features for detecting deceptive reviews. They reused datasets introduced in~\cite{2011Ott} and~\cite{Ott2013}, which contained $1,600$ hotel reviews. They used two types of features: 1) 77 lexical features, including 46 ``character-based'' features, such as character count and occurrences of special characters, and 31 ``word-based'' features, such as token counts and average token lengths, and 2) 157 syntactic features, including 7 ``occurrences of punctuation'', such as ``!'' and ``@'', and 150 ``occurrences of function words.'' 
To conduct their experiment, they examined polynomial SVM and Sequential Minimal Optimization (SMO) and Naive Bayes using the WEKA tool~\cite{WEKA2009}. They also performed their models on lexical, syntactic and combined features separately and achieved the best F-measures 81\%, 76\% and 84\% respectively by using SMO.

Li et al.~\cite{2014li} investigated cross-domain evaluation for detecting fake reviews. They built their dataset containing three domains: Hotels, Restaurants, and Doctors. To create their fake reviews, they employed Turkers and domain-experts; whereas, for the real reviews, customer reviews were used. Three groups of features were used in their experiments: uni-gram, LIWC, and POS. First, they performed a classification task within each domain by using SAGE (Sparse Additive Generative Model). Their best accuracy in each domain (i.e., Hotel, Restaurants, and Doctors) was around 81.8\%, 81.7\% and 74.5\%, respectively, using uni-gram features. In order to conduct a  cross-domain classification, they trained SAGE and SVM classifiers on the Hotel dataset and tested the trained model on other domains. The best accuracy achieved was 78.5\% by the SVM model on the Restaurant dataset using uni-gram features. The trained models did not perform well on the Doctor dataset. The best accuracy was around 64\% for the SAGE model using LIWC features. Having accomplished these experiments, they reported several general rules about fake reviews.

\section{Technical Background}
\label{sec:background}

\subsection{Machine/Deep Learning Classifiers}

We conducted our experiments using seven classifiers: Decision Tree (DTC), Random Forest (RF), Support Vector Machine (SVM), Extreme Gradient-Boosting Trees (XGBT), Multilayer Perceptron (MLP), Logistic Regression (LR), and Naive Bayes (NB). We also implemented the linguistic features using the spaCy 2.2.3 library.



\subsection{Feature Importance in Random Forest}

In Random Forest, a single tree of the ensemble is built using $N$ samples. Every node $t$ of the tree contains $N_t$ samples and is assigned a split $s_t$, generating two children $t_L$ and $t_R$, containing $N_{t_L}$ and $N_{t_R}$ samples, respectively. The way that $s_t$ is determined is by maximizing Equation \ref{eq:imp-dec}, which denotes a decrease of an impurity measure $i(t)$ (e.g., Gini impurity or entropy in this work), where $p_L = N_{t_L}/N$, and $p_R = N_{t_R}/N$.

\begin{equation}
    \label{eq:imp-dec}
    \Delta i(s, t) = i(t) - p_L i(t_L) - p_R i(t_R)
\end{equation}


\begin{equation}
    \label{eq:imp-f}
    Imp(f) = \frac{1}{N_T} \sum_{T} \sum_{t \in T: v(s_t) = f} p(t) \Delta i(s_t, t)
\end{equation}

The importance of $f$ (i.e., $Imp(f)$) in Random Forest is calculated as the summation of the impurity decreases $\Delta i(s_t, t)$ weighted by $p(t)$, for all nodes $t$, where $p(t) = N_t/N$ and $v(t)$ is the feature used in the split $s_t$. Finally, this value is averaged over the $N_T$  trees in the ensemble \cite{louppe2013understanding}.


\subsection{Recursive Feature Elimination (RFE)}

Recursive Feature Elimination (RFE) \cite{guyon2002gene} is a procedure that addresses the problem of feature selection in three steps:

\begin{enumerate}
    \item Fit the model (classifier) to the dataset,
    \item Rank the features according to their weights, and 
    \item Remove the feature with the lowest weight.
\end{enumerate}

The steps are repeated until the remaining number of features is equal to the required number of features. As this algorithm uses a classifier to perform feature selection, it is considered as a wrapper method. 
We used the logistic regression classifier in RFE. During the fitting stage, we applied 5-fold cross-validation in order to find the optimum value for the regularization hyperparameter $C$, with values ranging from $10^{-4}$ to $10^{4}$ in a logarithmic scale. 
We normalized the values for each feature so the order of magnitude of the values does not interfere with the weights calculated.

\subsection{Kernel Density Estimation}

Kernel Density Estimation (KDE) is a non-parametric method to estimate the underlying probability density function (PDF) of observed data. Let $f$ be the true underlying PDF of the data and $\hat{f}$ the estimated PDF using KDE, then
If $\hat{f}$ is the estimated PDF using KDE, then

\[
    \hat{f}(x;h) = \frac{1}{Nh} \sum_{i=1}^{N} K(\frac{x - x_i}{h}),
\]

\noindent where $N$ is the number of samples, $h$ is a positive number usually called \textit{bandwith} or \textit{window width}, and $K$ is a kernel function \cite{wand1994kernel}. A common choice for the kernel is the Gaussian kernel, which we use in our experiments.

\subsection{Overlapping Coefficients (OVL)}

In this work, we utilized the overlapping coefficient (OVL) \cite{inman1989overlapping} to measure the similarity between two probability density functions. The hypothesis behind this is that for two probability density functions $pdf_{fake}$ and $pdf_{real}$ that were estimated with KDE using a feature $f$ with its values for fake and real samples, respectively, their OVL will be low in case that $f$ is effectively discerning between the two classes. 

The OVL for two generic probability density functions $f(x)$ and $g(x)$ is defined in Equation \ref{eq:ovl}, where $R_n$ is the $n$-dimensional space of real numbers and $min$ is a function that returns the minimum value of two PDFs evaluated in a specific $x$ value. As each feature in our dataset is one-dimensional, the value for $n$ will be  $1$ in our case ($n = 1$).

\begin{equation}
    \label{eq:ovl}
    OVL = \int_{R_n} min(f(x), g(x)) dx
\end{equation}

\subsection{Relevant Features and Boruta}

Boruta is an algorithm that finds all the relevant features in a dataset \cite{kursa2010feature}. It is a wrapper around a Random Forest algorithm that determines the relevance of features through the comparison of them with their random versions. In addition to the initial set of features, Boruta computes a set of \textit{shadow} features corresponding to a random permutation of each original feature and appends it to the initial dataset. Afterwards, it fits a Random Forest to the augmented dataset and calculates the feature importance for the original and \textit{shadow} features. Next, the maximum importance among the shadow features is set as a threshold to compare the importance of regular features. If an importance of a normal feature is greater than this threshold, then the corresponding feature gets a \textit{hit}, otherwise it gets a \textit{no hit}. After a fixed number of iterations, each regular feature will contain a series of binary values (\textit{hit} or \textit{no hit}) that can be compared against the probability mass function of a binomial distribution with $p = 0.5$ and $n = $ number of iterations. Having the binomial probability mass function, Boruta establishes three areas: not relevant, tentative, and relevant. Finally, a feature is deemed relevant, not relevant, or inconclusive depending on the section that is projected in the probability mass function.

\section{Experimental Setup}
\label{sec:setup}

\subsection{Dataset: The ``Restaurant Dataset''}
We created our own dataset for this experiment\footnote{
https://github.com/asiamina/FakeReviews-RestaurantDataset}. Our dataset, called the ``Restaurant Dataset", was initially used in our previous work \cite{DADA2020-Deception}. The dataset consists of reviews about three local restaurants. For real reviews, we extracted reviews from legitimate users about those restaurants from online resources. For the fake reviews, four undergraduate students wrote imaginary fake reviews for the restaurants. The dataset is arranged in a way to have equal numbers of positive and negative reviews and also equal numbers of fake and real reviews (110 reviews in total).

\subsection{Classification Metrics}

\begin{gather*}
    Accuracy = \frac{TP + TN}{TP + TN + FP + FN}, \\
    Precision = \frac{TP}{TP + FP}, \\
    Recall = \frac{TP}{TP + FN}, \\
    F_1 = \frac{2 \times Precision \times Recall}{Precision + Recall}.
\end{gather*}

An Accuracy of $1.0$ indicates that the predicted labels and observed labels are identical. Precision is the ratio of true positives against all predicted positive labels. Recall is the ratio of predicted true positives against all observed positive labels. The $F_1$ score is the harmonic mean of Precision and Recall, which yields a more informative performance metric when Precision and Recall have dissimilar values. Although we report accuracy and $F_1$, we determine overall model performance using accuracy, because our balanced dataset prevents accuracy to report misleading values.

\subsection{Hyperparameter Tuning and Cross-Validation}

We report the accuracy and $F_1$ scores for the classifiers using 10-fold cross-validation. Furthermore, we performed grid searches to determine the best set of hyperparameters for the classifiers every time a new subset of selected features was used. Because of this, the hyperparameters vary according to the features with which the classifiers were fitted.

\section{Results}
\label{sec:results}

\subsection{Overlapping Probability Density Functions}

\begin{figure*}[hbt!]
  \centering
  \begin{subfigure}{.33\textwidth}
      \center
      \includegraphics[width=\linewidth]{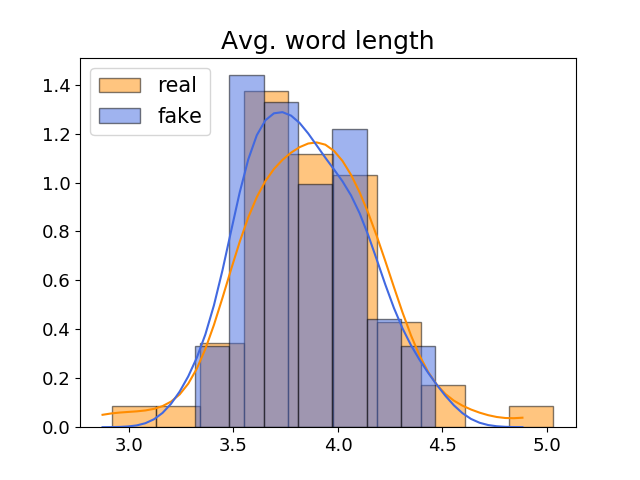}
      \caption{}
      \label{}
    \end{subfigure}%
       \begin{subfigure}{.33\textwidth}
      \centering
      \includegraphics[width=\linewidth]{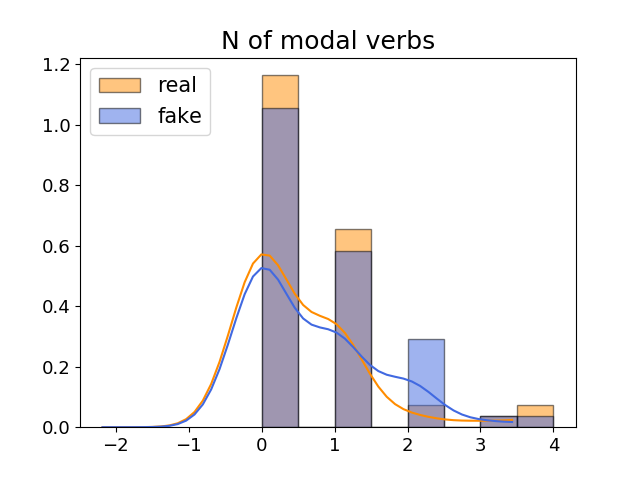}
      \caption{}
      \label{}
    \end{subfigure}%
      \begin{subfigure}{.33\textwidth}
      \centering
      \includegraphics[width=\linewidth]{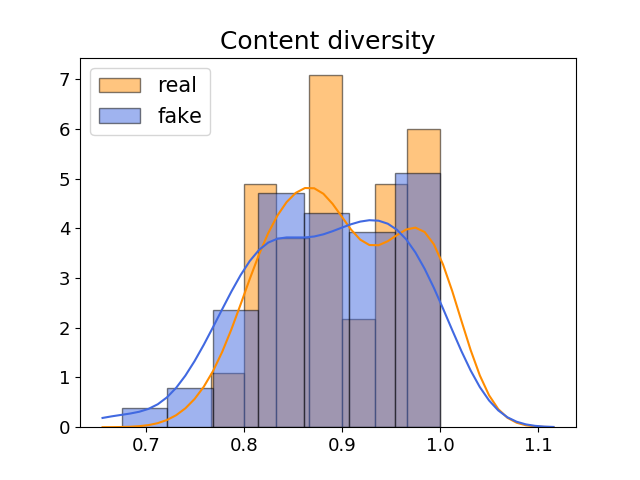}
      \caption{}
      \label{}
    \end{subfigure}%

     \begin{subfigure}{.33\textwidth}
      \centering
      \includegraphics[width=\linewidth]{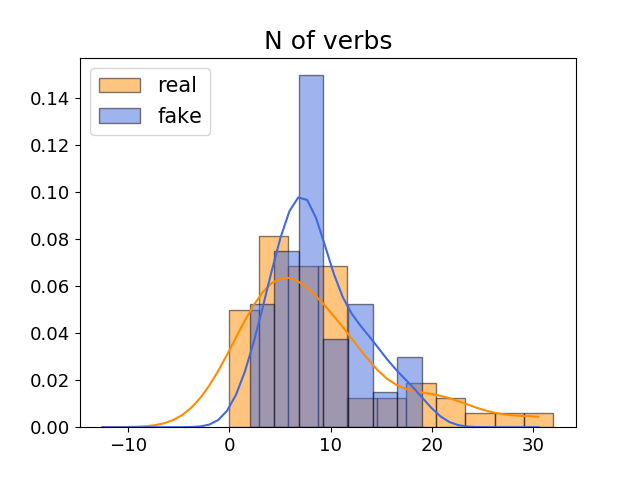}
      \caption{}
      \label{}
    \end{subfigure}%
    \begin{subfigure}{.33\textwidth}
      \centering
      \includegraphics[width=\linewidth]{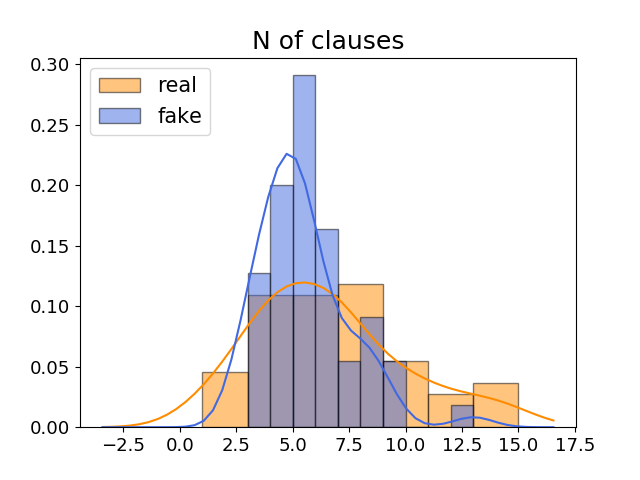}
      \caption{}
      \label{}
    \end{subfigure}%
    \begin{subfigure}{.33\textwidth}
      \centering
      \includegraphics[width=\linewidth]{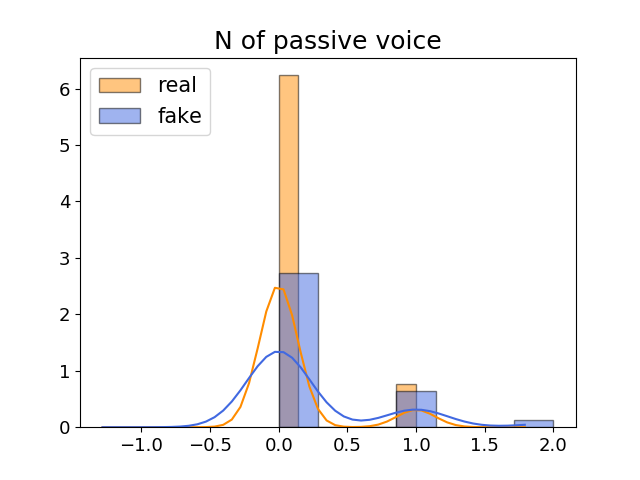}
      \caption{}
      \label{}
    \end{subfigure}%

    \begin{subfigure}{.33\textwidth}
      \centering
      \includegraphics[width=\linewidth]{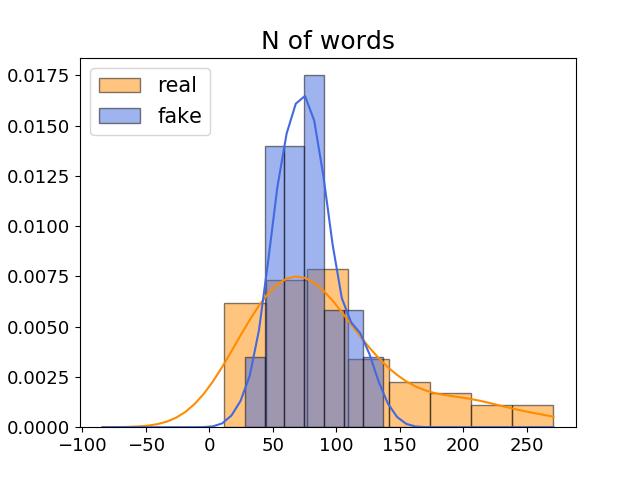}
      \caption{}
      \label{}
    \end{subfigure}%
    \begin{subfigure}{.33\textwidth}
      \centering
      \includegraphics[width=\linewidth]{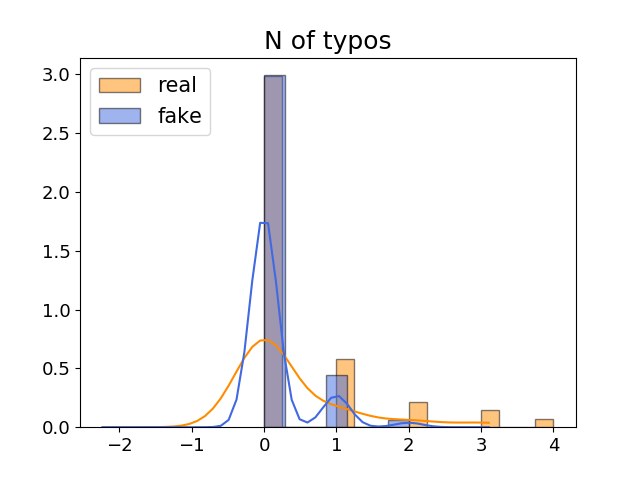}
      \caption{}
      \label{}
    \end{subfigure}  
    \begin{subfigure}{.33\textwidth}
      \centering
      \includegraphics[width=\linewidth]{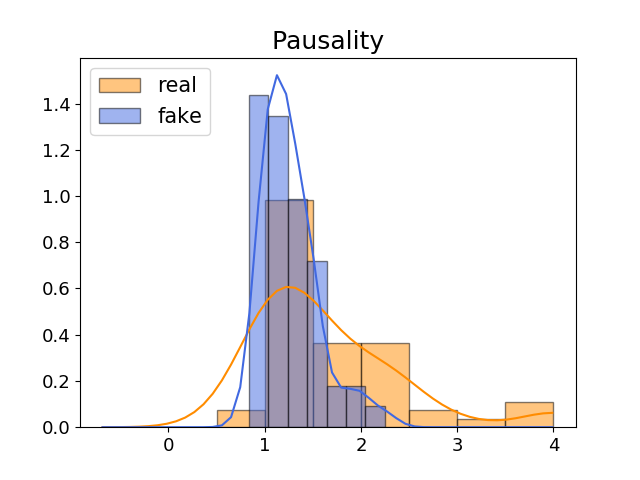}
      \caption{}
      \label{}
    \end{subfigure}%
 \caption{Histograms and PDFs for selected features.}
  \label{fig:hists}
\end{figure*}

Figure \ref{fig:hists} shows the histograms for a selected set of features, along with the probability density functions (PDFs) for both real and fake reviews. The histograms are displayed as probability density functions, and not as raw counts for each bin. In addition, we performed KDE using a Gaussian kernel to approximate the smoothing PDF for each feature separated by class (i.e., fake and real).
The histograms are ordered from the highly overlapped features for fake and real reviews to the features with medium level of overlaps.

Table \ref{tab:feat-overlaps} reports the overlapping coefficients for the probability density functions using KDE for each feature. The overlap coefficient scales are grouped into three classes: 1) very high overlap, 2) high overlap, and 3) medium overlap. As the table shows, pausality was identified as the least overlapped feature between fake and real reviews; whereas, average word length is the feature highest overlap between fake and real reviews. 

\begin{table}[h!]
    \centering
    \caption{Ordered overlapping coefficients.}
    \begin{tabular}{|l|r|c|}
        \hline
                              \multicolumn{1}{|c|}{\bf Feature} &   \multicolumn{2}{c|}{\bf Overlapping } \\
                              \cline{2-3}
                      \multicolumn{1}{|c|}{\bf } &   \multicolumn{1}{c|}{\bf  Coefficient} &
                     \multicolumn{1}{c|}{\bf  Scale} \\
        \hline
                    Pausality &  0.585425 & Medium \\  
                   N. of typos &  0.630281 & Medium \\
                   N. of words &  0.646805  & Medium \\ 
              N. of adjectives &  0.684258  & Medium \\
              \hline
                   Redundancy &  0.701065  & high \\    
         Avg. sentence length &  0.722965  & high \\
           N. of passive voice &  0.726593 & high  \\    
                 N. of clauses &  0.726634  & high \\
                   N. of verbs &  0.759618  & high \\   
            Lexical diversity &  0.791875 & high \\
            \hline
                Avg. NP length &  0.831418 & Very High \\
                  Emotiveness &  0.872769 & Very High \\
            Content diversity &  0.895416 & Very High \\ 
             N. of modal verbs &  0.906333 & Very High \\ 
              Avg. word length &  0.911932 & Very High \\
        \hline
    \end{tabular}
    \label{tab:feat-overlaps}
\end{table}

\subsection{Feature Significance}

The second column in Table \ref{tab:feats_rf} shows the normalized feature importance in descending order for each feature after fitting a random forest, an ensemble learning technique, to our dataset. We applied random forest classifiers for the real and fake reviews using the Python library of  
{\tt RandomForestClassifier()}. We then used the sklearn instances of the models using the {\tt .feature\_importances\_} attribute, which returns an array of each feature's significance in determining the splits. We then took an average for each feature's significance to measure the overall importance level. More specifically, the final feature importance is calculated averaging the feature importance of each tree, whereas the importance of a certain feature in a tree is calculated as its mean decrease impurity \cite{breiman2001random}.

\begin{table}[h!]
    \centering
    \caption{Random Forest feature significance, iteration of first appearance, and number of hits per each feature.}
    \label{tab:feats_rf}
    \begin{tabular}{|l|r|c|c|c|}
                       \hline
                              \multicolumn{1}{|c|}{\bf Feature} &   \multicolumn{1}{c|}{\bf RF Feature } &   \multicolumn{1}{c|}{\bf Appears on } & \multicolumn{2}{c|}{\bf  Boruta Output} \\
                              \cline{4-5}
                      \multicolumn{1}{|c|}{\bf } &   \multicolumn{1}{c|}{\bf  Significance} &
                     \multicolumn{1}{c|}{\bf  Iteration \#} & \multicolumn{1}{c|}{\bf Hits} & \multicolumn{1}{c|}{\bf Scale} \\
        \hline
               Redundancy &  0.132643 & 3 & 98 & High\\
                Pausality &  0.119000 & 2 & 98 & High\\
     Avg. sentence length &  0.103888 & 14 & 91 & High\\
          N of adjectives &  0.100279 & 1 & 93 & High\\
               N of words &  0.092147 & 5 & 89 & High\\
               \hline
        Lexical diversity &  0.077524 & 4 & 60 & Medium\\
                \hline
           Avg. NP length &  0.061619 & 8 & 43 & Low\\
               N of verbs &  0.061500 & 15 & 37 & Low\\
             N of clauses &  0.059991 & 7 & 26 & Low\\
              Emotiveness &  0.058063 & 11 & 2 & Low\\
        Content diversity &  0.052171 & 10 & 0 & Low\\
         Avg. word length &  0.045357 & 13 & 0 & Low\\
               N of typos &  0.015418 & 9 & 0 & Low\\
         N of modal verbs &  0.011928 & 12 & 0 & Low\\
       N of passive voice &  0.008471 & 6 & 0 & Low\\  
    \hline
    \end{tabular}
\end{table}

Since we do not have any priori knowledge about the most important features, we perform RFE to get feature sets of size one up to fifteen, repeatedly. Let $F_i$ be the set of selected features at iteration $i$, and $f_i$ a singleton containing the feature that appears for the first time in $F_i$ at 
iteration $i$. Note that $size(F_i) = i$ for all $i$, and $F_i \subset F_j$ if $i < j$. $F_i$ can be obtained recursively using Equation \ref{eq:rfe-set}. 

\begin{equation}
    \label{eq:rfe-set}
    \begin{split}
        F_0 & = \O \\
        F_i & = f_i \cup F_{i - 1}
    \end{split}
\end{equation}

The third column in Table \ref{tab:feats_rf} contains the iteration in which each feature appears for the first time, i.e., the $i$'s for the $f_i$'s. For example, from this column, the features selected using RFE after five iterations are the {\it 1) number of adjectives}, 2) {\it pausality}, 3) {\it redundancy}, 4) {\it lexical diversity}, and 5) {\it number of words}.


Table \ref{tab:num-acc-f1} shows the accuracy and $F_1$ achieved by the classifier for each subset of selected features. As Table \ref{tab:num-acc-f1} shows, the accuracy fluctuates between $0.6$ and $0.8$ for all classifiers. The $F_1$ scores are around $0.7$ for most of the classifiers. As indicated in Table \ref{tab:rfe-acc-f1}, the best values for accuracy and F scores are achieved when the number of features considered in the classification is between $3$ and $9$. The MLP classifier outperforms the other classifiers with 73.45\% and 70.84\% for accuracy and $F_1$ scores on average, respectively.

Table \ref{tab:rfe-acc-f1} contains a summary with the highest accuracy and $F_1$ score, alongside with the number of features with which these metrics were obtained (i.e., $size(F_i)$). MLP reports the highest accuracy and $F_1$ among the classifiers, with 79.09\% and 76.98\%, respectively. These values are obtained with only four features: {\it 1) number of adjectives}, 2) {\it pausality}, 3) {\it redundancy} and 4) {\it lexical diversity}. In general, the number of features for the maximum accuracy coincides with that of the maximum $F_1$ score, with the only exception of DTC. Compared to the efficiency of \cite{DADA2020-Deception} also shown in Table \ref{tab:num-acc-f1}, we achieved higher accuracy and $F_1$ score in our experiments. This suggests that, when compared to document embedding (Doc2Vec), the linguistic features employed in this work are more useful with regards to successfully discerning between fake and trustworthy reviews. 

\begin{table}[h!]
    \centering
    \caption{Maximum accuracy and $F_1$ for the classifiers.}
    \begin{tabular}{|l|r|c|r|c|}
        \hline
                               & \multicolumn{2}{c|}{\bf Accuracy} &
                              \multicolumn{2}{c|}{\bf $F_1$} 
                              \\
                              \cline{2-5} 
                              & & \multicolumn{1}{c|}{\bf \# of} & & \multicolumn{1}{c|}{\bf \# of}\\
                              \multicolumn{1}{|c|}{\bf Classifier}&\multicolumn{1}{c|}{\bf Max }& \multicolumn{1}{c|}{\bf features} &\multicolumn{1}{c|}{\bf Max }& \multicolumn{1}{c|}{\bf features} \\
                     \hline
                   SVM &  77.27\% & 3 & 76.12\% & 3 \\  
                   NB &  73.63\% & 4 & 66.04\% & 4 \\
                   RF &  75.45\%  & 3 & 73.67\% & 3 \\ 
                   LR &  73.63\%  & 7 & 70.39\% & 7 \\
                   DTC &  73.63\%  & 3 & 70.31\% & 9 \\    
                   XGBT &  74.54\%  & 6 & 71.76\% & 6 \\
                   MLP &  79.09\% & 4 & 76.98\% & 4 \\
        \hline
    \end{tabular}
    \label{tab:rfe-acc-f1}
\end{table}

\subsection{Boruta Feature Relevances}
The fourth column in Table \ref{tab:feats_rf} shows the number of {\it hits} of each feature after applying Boruta with 100 iterations. Under this setup, the relevant features turned out to be {\it pausality}, {\it average sentence length}, {\it number of words}, {\it number of adjectives}, and {\it redundancy}; whereas, only the lexical diversity was determined as tentative, leaving out the {\it average noun-phrase length}, {\it number of verbs}, {\it number of clauses}, {\it emotiveness}, {\it content diversity}, {\it average word length}, {\it number of typos}, {\it number of modal verbs}, and {\it number of uses of passive voice} as irrelevant features. These results agree with the features with higher importance as listed in Table \ref{tab:feats_rf}, where the first five features coincide with the relevant ones determined by Boruta. Note that {\it lexical diversity} ranked sixth in Table \ref{tab:feats_rf}.

\begin{table*}[h!]
    \centering
    \caption{A comparison of the accuracy and F scores.}
    \begin{tabular}{|c|lllllll|lllllll|}
    \hline
    \bf \# of & \multicolumn{7}{c|}{\bf Accuracy\%} & \multicolumn{7}{c|}{\bf $F_1$\%}\\
    \cline{2-15}
    \bf Features & \bf SVM & \bf NB & \bf RF & \bf LR & \bf DTC & \bf XGBT & \bf MLP & \bf SVM & \bf NB & \bf RF & \bf LR & \bf DTC & \bf XGBT & \bf MLP \\
    \hline
    1  &  66.36 &   60.0 &   60.0 &  65.45 &  66.36 &  66.36 &  65.45 &  54.21 &  41.86 &  48.39 &  58.98 &  51.91 &  54.21 &  60.37 \\
    2  &   70.0 &  66.36 &   70.0 &  63.64 &  64.55 &  69.09 &  70.91 &  63.72 &  55.33 &  63.53 &  56.33 &  55.77 &  62.85 &  65.26 \\
    \hline
    3  &  {\bf 77.27} &  69.09 &  {\bf 75.45} &  72.73 &  {\bf 73.64} &  71.82 &  77.27 &  {\bf 76.12} &  59.11 &  {\bf 73.67} &  {\bf 70.33} &   68.3 &  67.56 &  74.62 \\
    4  &  74.55 &  {\bf 73.64} &  72.73 &  71.82 &   70.0 &  71.82 &  {\bf 79.09} &  72.41 &  {\bf 66.04} &  70.94 &   70.2 &  64.92 &  69.09 &  {\bf 76.99} \\
    5  &  75.45 &   70.0 &  69.09 &   70.0 &  {\bf 73.64} &  71.82 &  72.73 &  72.29 &  60.67 &   66.7 &  66.94 &  70.19 &  70.02 &  71.02 \\
    6  &  72.73 &   70.0 &  71.82 &  71.82 &   70.0 &  {\bf 74.55} &  77.27 &  68.15 &  60.94 &   69.0 &  68.03 &  64.66 &  {\bf 71.76} &  76.57 \\
    7  &  74.55 &  69.09 &   70.0 &  {\bf 73.64} &   70.0 &  71.82 &  69.09 &  71.64 &  58.68 &  67.52 &  70.39 &  69.09 &  69.06 &  65.79 \\
    8  &  71.82 &  68.18 &   70.0 &  {\bf 73.64} &  71.82 &  70.91 &  77.27 &  68.44 &  59.66 &  67.29 &  69.85 &  68.71 &  68.49 &  75.65 \\
    9  &  71.82 &  68.18 &  69.09 &  72.73 &  72.73 &  67.27 &   70.0 &  67.98 &  57.75 &  66.71 &  69.58 &  {\bf 70.32} &  64.86 &  70.36 \\
    \hline
    10  &  73.64 &  68.18 &  70.91 &  72.73 &  67.27 &  69.09 &  74.55 &  74.47 &  56.89 &  68.59 &   69.0 &  61.22 &  64.53 &  72.23 \\
    11 &  72.73 &  68.18 &  72.73 &  72.73 &   70.0 &  72.73 &  78.18 &  68.65 &  56.08 &  70.32 &  68.78 &  66.95 &  69.08 &  74.96 \\
    12 &  71.82 &  68.18 &  74.55 &  68.18 &  70.91 &   70.0 &  71.82 &  68.55 &  56.08 &  71.45 &  63.08 &  68.69 &  67.94 &  67.02 \\
    13 &  69.09 &  69.09 &  71.82 &  67.27 &  66.36 &  73.64 &  70.91 &  65.32 &  58.46 &  70.08 &  64.22 &  62.13 &  71.07 &  68.88 \\
    14 &  72.73 &  68.18 &   70.0 &  66.36 &   70.0 &  73.64 &  73.64 &  69.53 &  57.08 &  66.98 &  63.64 &  67.51 &  71.07 &  70.16 \\
    15 &  70.91 &  68.18 &  69.09 &  66.36 &  71.82 &   70.0 &  73.64 &  66.62 &  56.05 &  66.01 &  63.64 &  69.29 &  68.35 &  72.86 \\
    \hline
    Average & 72.29 & 68.30 & 70.55 & 69.94 &69.94 & 70.97 & {\bf 73.45} & 68.54 & 57.37 & 67.14 & 66.19 & 65.31 & 67.32 & {\bf 70.84} \\
    \hline
    \hline
    doc2vec \cite{DADA2020-Deception} & 59.1 &  -- & {\bf 68.2} & -- & 54.5 & 63.6 & {\bf 68.2} & 60.8 & -- & 66.6 & -- & 50.0 & 55.5 & {\bf 69.5} \\
    \hline
    \end{tabular}
    \label{tab:num-acc-f1}
\end{table*}

\begin{table*}[!htb]
\centering
\caption{A comparison of the performance of our model with existing work.}
\begin{tabular}{|l|l|l|c|c|c|}
\hline
\multicolumn{1}{|c|}{\multirow{2}{*}{\bf Reference}} & \multicolumn{1}{c|}{\multirow{2}{*}{\bf Dataset}} & \multicolumn{1}{c|}{\multirow{2}{*}{\bf Features}} & \bf Models & \multicolumn{2}{c|}{\bf Performance} \\ \cline{4-6} 
\multicolumn{1}{|c|}{} & \multicolumn{1}{c|}{} & \multicolumn{1}{c|}{} & \bf & \bf ACC & $\bf F_1$ \\ \hline
Li et al.~\cite{FangtaoLi2011} & Amazon reviews about products & \begin{tabular}[c]{@{}l@{}}review/reviewer/product centric\end{tabular} & NB &  & 58\% \\ \hline
\multirow{2}{*}{Li et al. \cite{2014li}} & TripAdvisor reviews about hotels & \multirow{2}{*}{uni-gram} & \multirow{2}{*}{SAGE} & 81.8\% &  \\ \cline{2-2} \cline{5-6} 
 & Reviews about resturants &  &  & 81.7\% &  \\ \hline
Ott et al. \cite{2011Ott} & TripAdvisor reviews about hotels & LIWC+bi-gram & SVM & 90\% &  \\ \hline
Shojaee \& Murad \cite{2013Shojaee} & Reviews about hotels~\cite{2011Ott}+~\cite{Ott2013} & \begin{tabular}[c]{@{}l@{}}Writing-style   features  \\ (lexical+syntactic)\end{tabular} & SMO &  & 84\% \\ \hline
\multirow{4}{*}{Mukherjee et al. \cite{2013Mukherjee}} & \multirow{2}{*}{Yelp reviews about hotels} & uni-gram & \multirow{4}{*}{SVM} & 65\% & 69\% \\ \cline{3-3} \cline{5-6} 
 &  & Behavioral features + bi-gram &  & 84\% & 84\% \\ \cline{2-3} \cline{5-6} 
 & \multirow{2}{*}{Yelp reviews about hotels} & bi-gram+POS &  & 68\% & 68\% \\ \cline{3-3} \cline{5-6} 
 &  & Behavioral features + bi-gram &  & 86\% & 85\% \\ \hline
\end{tabular}
\label{tab:workcomp}
\end{table*}

\subsection{Common Significant Features}
\begin{figure*}[t]
  \centering
    \begin{subfigure}{.49\textwidth}
      \centering
      \includegraphics[width=\linewidth]{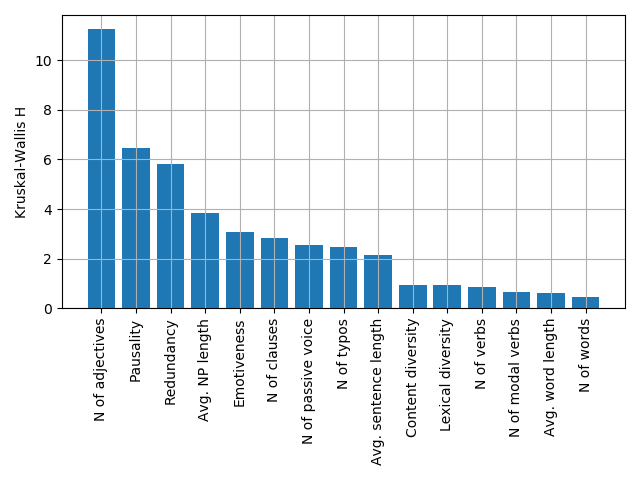}
      \caption{Kruskal-Wallis H score for each feature.}
      \label{fig:kw_scores}
    \end{subfigure}%
    \begin{subfigure}{.49\textwidth}
      \centering
        \includegraphics[width=\linewidth]{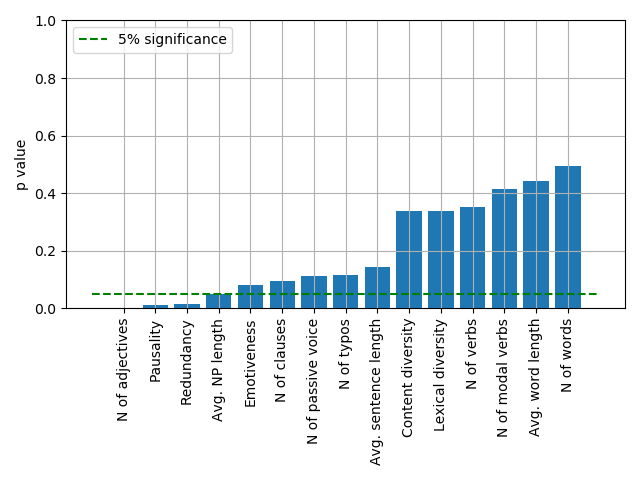}
        \caption{{\it p}-values of the Kruskal-Wallis ANOVA for each feature.}
        \label{fig:kw-pval}
    \end{subfigure}
    \caption{Anova and H scores.}
  \label{fig:mi_scores}
\end{figure*}

There exist some evident overlap among the first selected features by ANOVA, RFE, Boruta, and Random Forest. Figure \ref{fig:mi_scores} illustrates the magnitude of {\it p}-values for each feature using the non-parametric Kruskal-Wallis test. As the figure shows, the features that are strictly below the significance level of 5\% are 1) {\it the number of adjectives}, 2) {\it pausality}, and 3) {\it redundancy}, which are in compliance with the results shown in Table \ref{tab:feats_rf} (i.e., the five most important features), and thus deemed as relevant by Boruta. At the same time, these three features are the same as the ones selected in the third iteration by RFE. This presents the evidence that {\it the number of adjectives}, {\it pausality}, and {\it redundancy} are likely to be the key features in the context of discerning whether a review is fake or trustworthy.

\subsection{RF Feature Importance and OVL Coefficient}
In order to explore the relationship between the RF feature importance and its OVL, we computed the Spearman $\rho$ between these two series. The Spearman $\rho$ is a value that ranges from $-1$ to $1$ and denotes the degree of rank correlation between two variables
, where values close to $-1$ and $1$ denote a negative and positive rank correlation, respectively. The Spearman $\rho$ between the RF importance of features and their OVL is approximately $-0.52$ with a p-value of $0.04$. This shows a negative correlation between the OVL and the feature importance computed in RF, which supports the hypothesis that more important features have less overlap between their PDFs. However, Figure \ref{fig:hists} shows that {\it number of typos}, {\it number of passive voice}, and {\it number of modal verbs} are rather sparse features in our dataset. This sparsity can also be considered a cause for their low RF significance in Table \ref{tab:feats_rf}, as they are potentially noisy features. If we exclude these three features and calculate $\rho$ again, its new value is approximately $-0.853$, with a p-value of $0.0004$, which indicates a very strong negative linear relationship between RF feature importance and OVL coefficients.

\subsection{Feature Correlation and RF Feature Importance}

\begin{figure}[t]
  \centering
  \includegraphics[width=0.99\linewidth]{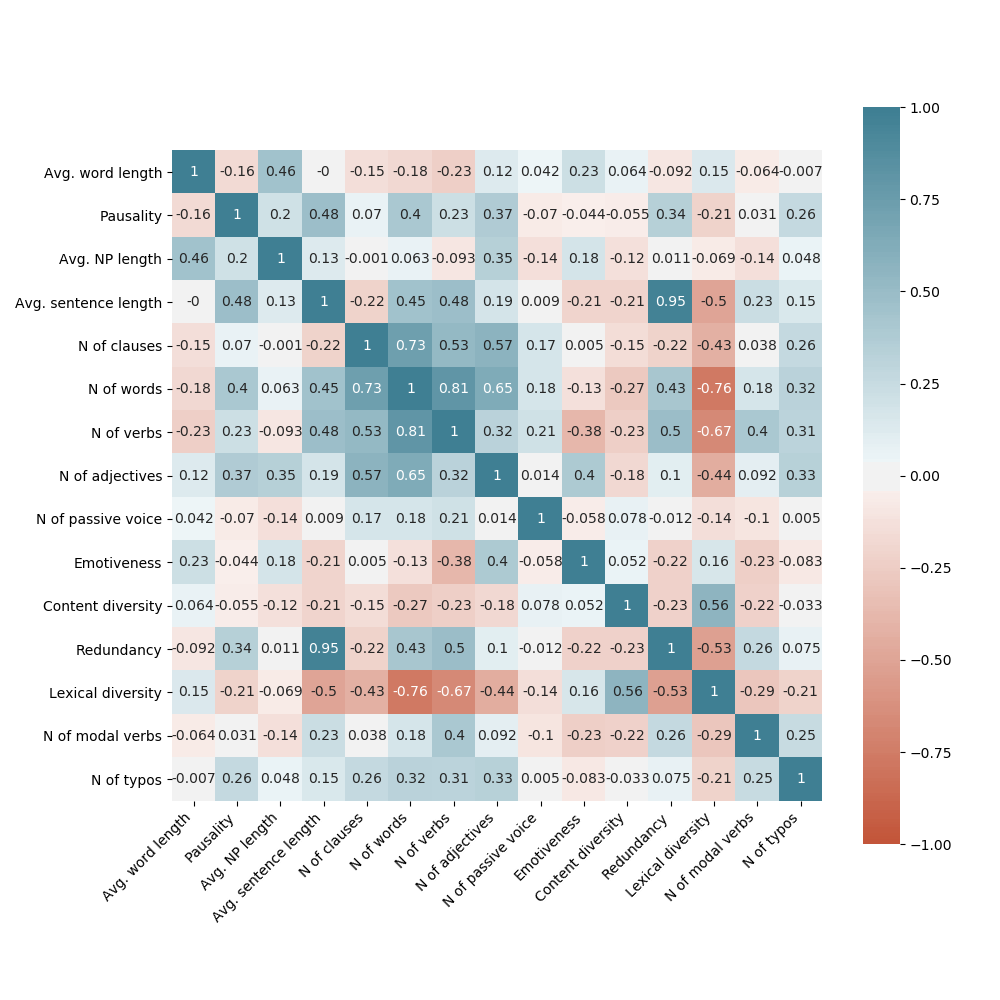}
  \caption{Spearman correlation (heat matrix) between features.}
  \label{fig:heatmap_corr}
\end{figure}

Figure \ref{fig:heatmap_corr} illustrates the pairwise Spearman correlation matrix of the features. The presence of average sentence length in the third place, over the number of adjectives (Table \ref{tab:feats_rf}) can be explained using the correlation among features. 
The {\it average sentence} length holds a very high correlation of $0.95$ with {\it redundancy}. 
In the context of Random Forest, as the features to split in each tree are a random subset of the initial set of features, it is possible that when the average sentence length is selected as a feature, the redundancy is not, and vice versa. Because they are so highly correlated, they will have a very similar importance in the trees where they are selected. 
If two features are selected at the same time in a tree, then it will depend on which one was used to split a node first, in the case of selecting redundancy first, the majority of impurity reduction will be attributed to redundancy, leaving a very low margin for the average sentence length to reduce the impurity and hence, being determined as an unimportant feature in the tree. The same analysis can show if the average sentence length is selected first. This behavior of interchangeable use of correlated features in random forest, in addition to other effects of correlated features, can be found in \cite{tolocsi2011classification}.

The high correlation between these features can also affect the interpretation of their weights in the logistic regression during RFE, as highly correlated features can make the model be numerically unstable with respect to the weights. This would explain why the average sentence length is included in the $14^{th}$ iteration, denoting as a very unimportant feature, whereas redundancy was selected early in the third iteration.

\section{Discussion}
\label{sec:discussion}

In this section, we compare the accuracy of our model with the existing work. Table \ref{tab:workcomp} provides the comparison. 

Mukherjee et al.~\cite{2013Mukherjee} discuss the reason for considerable gap between their results and Ott et al.~\cite{2011Ott} results, even though linguistic features were used in both cases. Mukherjee et al.~\cite{2013Mukherjee} investigated the word distribution in the fake reviews in both datasets and concluded that the reason is because of the differences in methods for extracting fake reviews. The fake reviews in Ott et al. \cite{2011Ott} dataset were written by Turkers, called ``pseudo fake''; whereas, Mukherjee et al.~\cite{2013Mukherjee} used fake reviews from real spammers filtered by Yelp.

Mukherjee et al.~\cite{2013Mukherjee} mentioned two main possibilities: 1) The Turkers did not try their best for crafting fake reviews, and 2) The spammers on Yelp exaggerated about their fakes. This observation implies the importance of the process of {\it ``fake review selection''} or ``{\it fake review creation.}''

Mukherjee et al.~\cite{2013Mukherjee} also state that since the fake reviews in ~\cite{FangtaoLi2011} and~\cite{jindal2007} were collected and labeled by human, their results are not comparable as it was shown that human judgment is around random guess for detecting spam reviews~\cite{2011Ott}. In addition, in their experiments they examined different types of features. Based on their results, using only linguistic features did not lead to a good performance (accuracy was between $55\%$ to $68\%$). However, incorporating additional behavioral features increased the performance significantly (around 20\% increase). Since the fake reviews in their dataset were those detected by Yelp, it can be a reason for similarity between their approach and Yelp's approach for spam review detection and not necessarily a reason for weakness of linguistic approaches.
It is apparent that extracting more features would increase the detection performance and accuracy. For example, Shojaee and Murad~\cite{2013Shojaee} achieved their highest performance using all 234 features. In our work, we used only 15 linguistic cues and achieved reasonable performance, which can be an indication for effectiveness of linguistic features. In addition, using different feature selection/reduction techniques, we could achieve 79\% accuracy with only four features. This implies that some linguistic features can have a great discriminatory nature for differentiating between fake and trustworthy reviews. 

\section{Conclusion and Future Work}
\label{sec:conclusion}

In this work, we explored the significance of 15 linguistic features towards the classification of written reviews as fake or trustworthy. Using RF feature importances, RFE, Boruta, and ANOVA, we were able to identify consistently the {\it number of adjectives}, {\it redundancy}, and {\it pausality} as the most important features in this task. We conducted our classification experiments using seven classifiers, where the MLP reports the highest accuracy with $79.09\%$ using only 4 features (i.e., the aforementioned ones plus {\it lexical diversity}). Having identified the most important features in the fake review detection task, we can apply these features in other similar subjects using transcripts or texts as their data. 

For future work, we are planning to apply these features to detect phishing attacks, which is a social engineering technique. Utilizing these features, we can classify phishing emails from benign emails automatically. The feature-based detection of fake reviews can be also applicable to similar problems such as malware detection \cite{DBLP:conf/bigdataconf/AbriSKSN19} that needs to be explored. It is also possible to reduce the number of linguistic features using deep learning techniques such as autoencoders and identify the most significant and influential features to build the model \cite{autoencoder2020}. An interesting approach to detecting fake reviews can be the adaptation of reinforcement learning to this problem where an agent learns the known and latent features that account for reviews being trustworthy or counterfeit reviews \cite{DBLP:conf/compsac/ChatterjeeN19}.

\section*{Acknowledgment}
This research work is supported by National Science Foundation under Grant No: 1723765.

\bibliographystyle{IEEEtran}
\bibliography{IEEEfull,sample-base}
\newpage

\end{document}